# Passive Non-line-of-sight Imaging for Moving Targets with an Event Camera

CONGHE WANG,[1,†] YUTONG HE,[2,†] XIA WANG,[1,3,*] HONGHAO HUANG,[2] CHANGDA YAN,[1] XIN ZHANG,[1] HONGWEI CHEN[2,4,*]

[1]Key Laboratory of Photoelectronic Imaging Technology and System of Ministry of Education of China, School of Optics and Photonics, Beijing Institute of Technology, Beijing 100081, China
[2]Beijing National Research Center for Information Science and Technology (BNRist), Department of Electronic Engineering, Tsinghua University, Beijing, 100084, China
[3]angelniuniu@bit.edu.cn
[4]chenhw@tsinghua.edu.cn.

Non-line-of-sight (NLOS) imaging is an emerging technique for detecting objects behind obstacles or around corners. Recent studies on passive NLOS mainly focus on steady-state measurement and reconstruction methods, which show limitations in recognition of moving targets. To the best of our knowledge, we propose a novel event-based passive NLOS imaging method. We acquire asynchronous event-based data which contains detailed dynamic information of the NLOS target, and efficiently ease the degradation of speckle caused by movement. Besides, we create the first event-based NLOS imaging dataset, NLOS-ES, and the event-based feature is extracted by time-surface representation. We compare the reconstructions through event-based data with frame-based data. The event-based method performs well on PSNR and LPIPS, which is 20% and 10% better than frame-based method, while the data volume takes only 2% of traditional method. © 2022 Optica Publishing Group

Non-line-of-sight (NLOS) imaging has attracted great attention with its widespread potential applications in object detection, autonomous driving, and anti-terrorist reconnaissance [1-3]. According to whether a controllable light source is used, NLOS imaging is classified as active NLOS imaging [4-5] and passive NLOS imaging [6-7].

Passive NLOS imaging shows promising application and research prospects due to its simple device and convenient acquisition. However, NLOS problem is known as an inverse problem in mathematics and we need to perform blind deconvolution, which is time consuming and computational burdened. Consequently, light-cone transform theory using matrix inverse [8], back projection algorithm based on photon time-of-flight information [9] and wave-based phasor field approach [10] are proposed successively. But few of these methods could be applied to passive NLOS because the steady-state detection mode of passive NLOS suffering from serious degradation of the speckle on the intermediary surface. Currently, speckle coherence restoration [11] and intensity based data driven reconstruction methods [12-13] are used to solve the ill posed passive NLOS imaging dilemma. Since the movement of target induce motion blur to the intensity speckle on the intermediary surface and superpose with the obscure caused by diffusion [14], most end-to-end deep-learning approach require extremely large training set [15] or perform well only on static NLOS targets [16]. In contrast, to realize high quality and efficient reconstruction with limited data set, we firstly put forward an approach for NLOS moving target reconstruction with an event camera. In this way, the dynamic information of the speckle intensity is precisely captured by the event detection paradigm.

Event camera [17], known as a novel neuromorphic vision device, only responses to brightness changes per-pixel asynchronously, while traditional frame-based cameras measure absolute brightness at a fixed rate. The record paradigm of event-based vision provides high temporal resolution, high dynamic range and low power consumption [18]. Therefore, it finds a large potential in challenging scenarios for standard camera, such as high speed, high dynamic range imaging [19] or object detection in slight change optical field.

Data collected by an event camera is recorded in the form of four Characteristic Parameters (CPs), time stamp $t$, spatial address $x$, $y$ and polarity flag $p$. The $i^{th}$ event is noted as $ev_i$ in Equation(1):

$$ev_i = [t_i, x_i, y_i, p_i]^T, i \in N. \qquad (1)$$

The feature representation of the event-based data mainly expressed by the four CPs. Event address map [20] and Time-surface [21] are two of typical methods. In our work, we adopt Time-surface method to represent the event-based data and extract featured speckles, which contain rich texture information of the target movement. A brief introduction of event-based vision and the detailed illustration of Time-surface representation method are given in the Supplement 1.

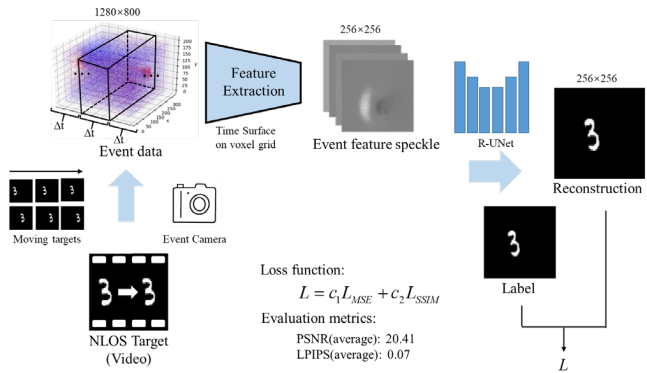

Fig.1. The flowchart of the proposed event embedded passive NLOS imaging.

To extract the dynamic information of intensity speckle movement, and reconstruct the hidden moving target, we put forward an event embedded data-driven method that fuses the event features with end-to-end deep learning approach. As shown in Figure 1, we display a video containing a parallel moving target with a smart phone, and use an event camera to record the dynamic speckle on the intermediary surface. We perform Time-surface calculation on voxel grid [21] to extract the featured speckle, and represent the asynchronous events in different time intervals by series of 2 dimensional intensity images. The Time-surface is expressed by Equation (2):

$$S_i(\rho, f) = e^{-(t_i - T_i(\rho, f))/\tau} \qquad (2)$$

$S_i$ is the Time-surface value of event $e_i$, which is defined by applying an exponential decay kernel with time constant $\tau$ on the values of context time stamp $T_i(\rho, f)$. And $T_i$ is defined to represent the time-context information around an incoming event $e_i$ as the array of most recent events times at $t_i$ for the neighborhood pixels in the adjacent area with radius of $\rho$:

$$T_i(\rho, f) = \max_{j \le i} \{t_j \mid r_j = r_i + \rho, f_j = f\} \qquad (3)$$

For speckle reconstruction, we employ UNet structure [22] and skip connection to perform multi-scale feature fusion and add residual block to avoid gradient explosion when training. Batch Normalization (BN) layers are induced to deeper the trainable depth and serve as regularity constraints, ensuring the network generalization ability. The loss function is made up of Mean-Square Error (MSE) and Structure Similarity Index Measure (SSIM). Peak Signal-to-Noise Ratio (PSNR) and Learned Perceptual Image Patch Similarity (LPIPS) with vgg model are used as evaluation metrics.

For the experimental proof of the event-based approach, we constructed experimental setups, as shown in Figure 2. In the experiment, we displayed a video which provided the self-luminous moving target in NLOS region of the event camera, blocked by the obstacle. The moving speckle on the intermediary surface is record by the event camera (CeleX-V). The detailed parameters are given in Supplement 2.

The targets used in the experiment contains characters of number digits selected from MNIST train set, MNIST test set and PRINT test set (Arial font numbers), with the size of $3cm \times 3cm$, which is placed $25cm$ away from the aluminum fender. We selected 14 different kinds of characters for each digit (0-9) in MNIST train set and test set, then acquire both event-based data and frame-based data with different mode of the CeleX-V camera. When recording the moving speckle in Full picture (F) mode, we get series of speckle screenshots in different position with the framerate of 100fps, while in Event Intensity (EI) mode, we get a stream of event-based data of the speckle movement. Data collected by these two modes are calibrated to the ground truth by the time-stamp and made into image format data sets NLOS Event Speckle (NLOS-ES) and NLOS Frame Speckle (NLOS-FS) correspondingly.

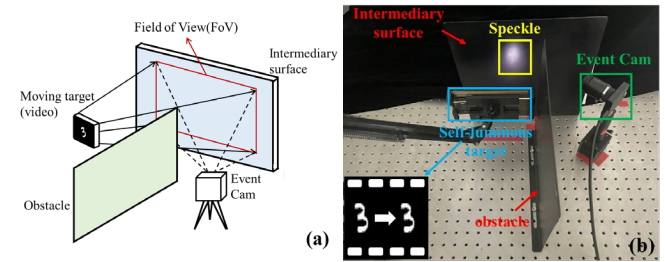

Fig. 2. The experimental setup of our event detecting NLOS: (a) basic principle of non-line-of-sight scene, (b) experimental settings.

To the best of our knowledge, we firstly establish the NLOS-ES data set, which contains 4080 images in train set and validation set, and 210 images in test set. The train set is made up of 3950 event feature Time-surface speckle of 130 targets (13 groups 0-9) at different position, while validation set contains 130 images. The test set consist of 110 images with 10 targets selected from MNIST test set and 100 images with 10 target digits (0-9) in Arial font. As the counterpart, NLOS-FS contains the corresponding frame-based intensity speckle with 4180 images in total.

We trained our Residual-UNet (R-UNet) on NLOS-ES train set by Adaptive Moment estimation (Adam) optimizer with Intel i9-7900X CPU and Nvidia RTX 3090 GPU for 800 epochs. To achieve fair comparisons, we trained the R-UNet on NLOS-FS train set with the same configuration [23]. The reconstruction result on validation set verified that the fully convolutional network has shown the ability of image generating and enhancement with limited data set. The structure of our R-UNet and training parameters are given in the Supplement 3.

The reconstruction quality of the moving target is assessed from two perspectives, the visual reconstruction quality and the position accuracy. For the former, we introduce peak signal-to-noise ratio (PSNR) and learned perceptual image patch similarity (LPIPS) to evaluate the reconstructions. The reconstruction accuracy of Frame-based method (F method) and Event-based method (E method) on MNIST test set and PRINT test set are shown and compared in Figure 3 and Figure 4 respectively. It is obvious that the proposed E method with event-based data shows much better reconstruction quality than F method, especially in recognizing the digits. The test results on MNIST test set and untrained print font digits prove the generalization ability of the network within certain extent. As for the position accuracy, we define the index Contour distance (Cd) to measure the position of the reconstructions. The Cd value is evaluated by the average distance between the left edge and the digit left contour (made up by first pixels with grayscale of 255 in each row after image binarization).

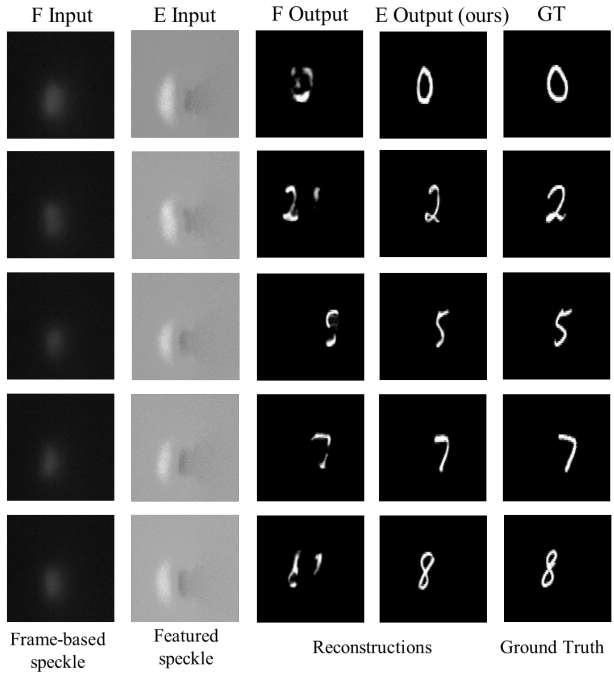

Fig. 3. Part of the reconstruction results for MNIST test set in both NLOS-ES and NLOS-FS.

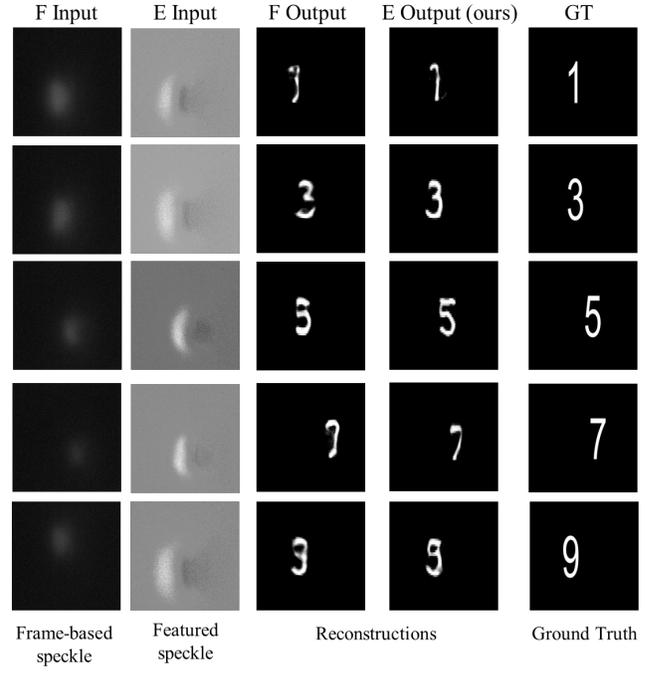

Fig. 4. Part of the reconstruction results for PRINT test set in both NLOS-ES and NLOS-FS.

Digit 7 in MNIST test set and digit 3 in PRINT test set are displayed as examples in this letter. As shown in Figure 5, the reconstruction by E method are closer to ground truth than that of F method. The average Cd deviation of E method is far smaller than that of F method, as shown in Figure 6.

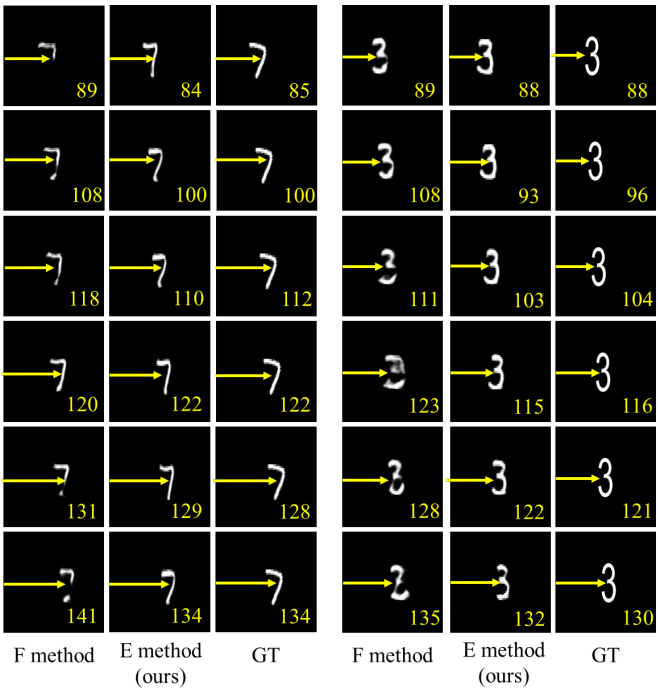

Fig. 5. Reconstructions of NLOS moving target at different position through event-based method and frame-based method. Six different positions of digit 7 (MNIST test set) and digit 3 (PRINT test set) are displayed as examples. The Cd value (pixel) is labeled at the corner of each frame.

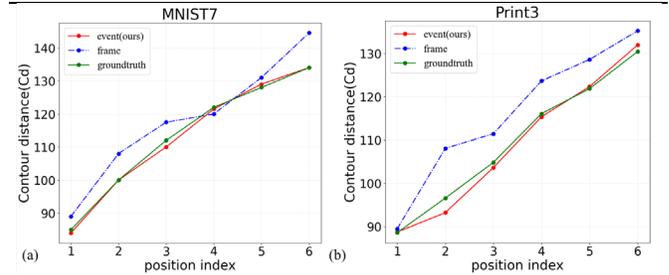

Fig. 6. The Cd value of NLOS reconstructions at different position. (a), (b) are the Cd values of reconstructions shown in Fig.5, digit 7 and digit 3, respectively.

From the other hand, the visual reconstruction accuracy of E method is also intuitively higher than F method. One could see from metrics LPIPS and PSNR shown in Figure 7, E method perform evidently better on both of two metrics, indicating that the event-based approach exceeds the frame-based method with a 100 fps traditional camera, under the same dataset size and network model. Besides, the data volume of the event-based data is only about 2% (5.96MB vs. 291.51MB) of that record by a frame camera.

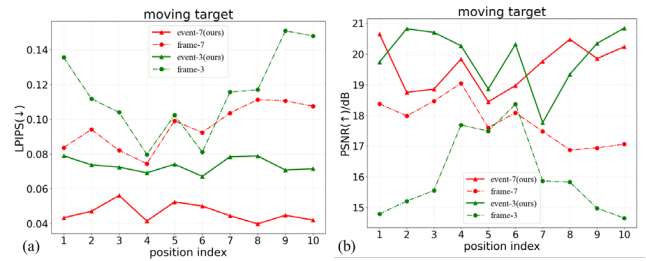

Fig. 7. The evaluation metrics of NLOS moving target reconstructions at different position in Fig.5. (a), (b) are the LPIPS and PSNR for digit 7 (MNIST test) and digit 3 (PRINT test) at ten different positions, respectively.

We statistically analyzed the reconstruction accuracy indexes of ten digits in both MNIST test set and PRINT test set at different positions. The average LPIPS and PSNR of each reconstructed frame for different test digits with event-based method and frame-based method are shown in Figure 8, respectively. One could see from Figure 8. (a), (b), the reconstructions of each digit by E method have improved about 25% and 10% on LPIPS than that by F method, respectively, which demonstrate the higher quality of human perception. As for PSNR, in Figure 8. (c), (d), E method scored higher than F method on every test targets.

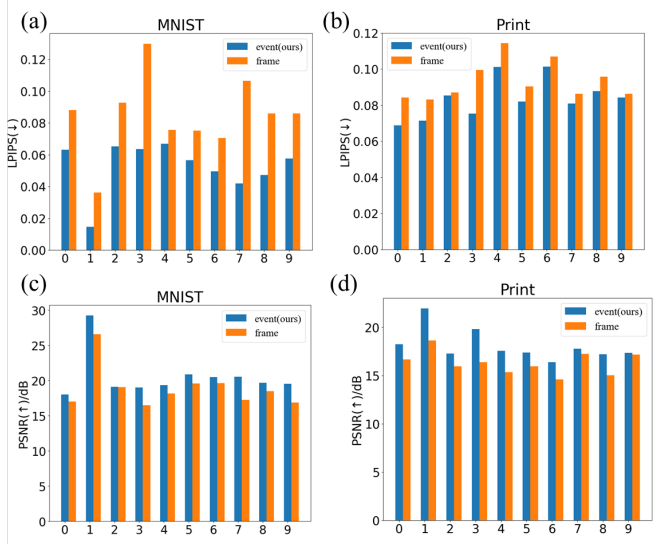

Fig. 8. The evaluation metrics of all reconstruction results with event-based method and frame-based method. (a), (c) are average LPIPS and PSNR of each reconstructed frame for ten digits in MNIST test set. (b), (d) are average LPIPS and PSNR of each reconstructed frame for ten print font digits in PRINT test set.

In summary, we firstly proposed a new detection and reconstruction method for passive NLOS imaging, using a novel detector: event camera. The event-based method which extracts rich dynamic information of the speckle movement and provide a physical foundation for data-driven passive NLOS imaging. Compared with deep learning approach with traditional camera, the event-based approach shows better performance when reconstructing NLOS moving targets. We carried out experiments on two types of targets with different distribution form, and verified that the reconstruction quality is significantly improved with the E method we addressed in both visual accuracy and position accuracy. The generalization ability on PRINT test set indicates that our method has extracted more meaningful information of moving speckle with event detection paradigm, compared with traditional frame-based detection. We believe that the R-UNet framework together with the NLOS-ES dataset is a big step and can inspire new ideas towards the development of feature embedded passive NLOS imaging with multi-detector information fusion [24], and NLOS target tracking [25]. Our event-based approach fuses the event paradigm information of NLOS moving target with end-to-end data driven methods, which reduces the data volume greatly, and has great potential to facilitate the further research of learned feature embedding for passive NLOS and its applications.

**Funding.** National Natural Science Foundation of China (62031018).

**Acknowledgments.** The authors thank Yuwei Zhao, Jiabi Zhao, Kai Ma for their assistance in experimental setup and Ruixu Geng, Zheng Huang for useful discussions concerning data driven NLOS imaging with novel imaging devices and sensors.

**Disclosures.** The authors declare no conflicts of interest.

**Data availability.** Data underlying the results presented in this paper are not publicly available at this time but may be obtained from the authors upon reasonable request.

**Supplemental document**. See Supplement 1,2,3 and Visualization 1 for supporting content.

† These authors contributed equally to this Letter.